\documentclass[conference]{IEEEtran}
\IEEEoverridecommandlockouts
\usepackage[nocompress]{cite}
\usepackage{amsmath,amssymb,amsfonts}
\usepackage{algorithmic}
\usepackage{graphicx}
\usepackage{textcomp}
\usepackage{xcolor}
\usepackage{amsmath}
\usepackage{url}
\usepackage{hyperref} 
\hypersetup{
colorlinks   = true,
citecolor    = blue,
urlcolor    =  blue
}
\usepackage[numbers,sort&compress]{natbib}

\newcommand{\footURL}[1]{\footnote{\url{#1}}}
\def\BibTeX{{\rm B\kern-.05em{\sc i\kern-.025em b}\kern-.08em
    T\kern-.1667em\lower.7ex\hbox{E}\kern-.125emX}}
\usepackage{caption}
\usepackage{subcaption}   

\begin{document}

\title{Selecting Seed Words for Wordle using Character Statistics}

\author{\IEEEauthorblockN{Nisansa de Silva}
\IEEEauthorblockA{\textit{Department of Computer Science \& Engineering} \\
\textit{University of Moratuwa}\\
Moratuwa, Sri Lanka \\
NisansaDdS@cse.mrt.ac.lk}
}

\maketitle

\begin{abstract}
Wordle, a word guessing game rose to global popularity in the January of 2022. The goal of the game is to guess a five-letter English word within six tries. Each try provides the player with hints by means of colour changing tiles which inform whether or not a given character is part of the solution as well as, in cases where it is part of the solution,  whether or not it is in the correct placement. Numerous attempts have been made to find the best starting word and best strategy to solve the daily wordle. This study uses character statistics of five-letter words to determine the best three starting words
\end{abstract}


\section{Introduction}
The web-based word game \textit{Wordle}~\cite{Wardle2022Wordle}, since its introduction in November of 2021 has captured the minds of the Internet. The game-play is rather simple. It gives players six attempts to guess a five letter word. At each attempt, hints are given to the player with coloured tiles. The allure of the game is one part due to the false scarcity it has created by limiting it to one puzzle per day. 

Given the popularity of the game~\cite{anderson2022finding}, there have been a number of attempts to find the best first word for fast and accurate solving of the daily puzzle. The work by~\citet{Kandabada2022Wordle} suggested manually selected four words,~\texttt{[SPORT, CHEWY, ADMIX, FLUNK]}, as the best starting words. The work by~\citet{Sidhu2022Wordle} attempted to find the best starting word from a linguistic perspective while the work by~\citet{Horstmeyer2022Want} suggested to do the same by running over 1 million simulations.~\citet{Bram2022Two} have proposed two strategies to employ in winning the game which they derived from their experience with crosswords. Interestingly, there also have been studies done on the worst word with which to start the wordle~\cite{Butterfield2022Science}.~\citet{anderson2022finding} have used machine learning to find the optimal human strategy for solving the \textit{wordle}. 

\newcommand{\obj}{to derive the set of 3 optimum starting words for \textit{wordle} covering 15 different characters and ordered in the descending order of significance}

The objective of this work is~\obj. The rest of the paper is organised as follows, Section~\ref{sec:wordle} provides a brief introduction to \textit{Wordle}, Section~\ref{sec:method} describes our methodology, Section~\ref{sec:ExpRes} reports our experiments and results. Finally, Section~\ref{sec:Conlu} concludes the paper.

\section{Wordle}
\label{sec:wordle}
\begin{figure}[!hbt]
     \begin{subfigure}[!hbt]{0.24\textwidth}
         \centering      \includegraphics[width=\textwidth]{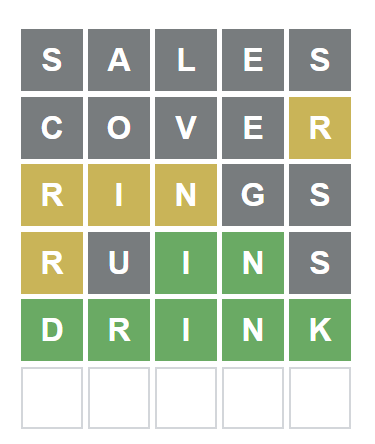}
         \caption{Guessing the word \textit{Drink}}       \label{fig:wordleDrink}
    \end{subfigure}
    \begin{subfigure}[!hbt]{0.24\textwidth}
         \centering      \includegraphics[width=\textwidth]{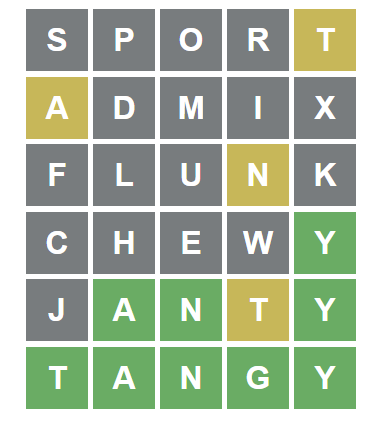}
         \caption{Guessing the word \textit{Tangy}}       \label{fig:wordleFavor}
    \end{subfigure}
    
     \begin{subfigure}[!hbt]{0.24\textwidth}
         \centering      \includegraphics[width=\textwidth]{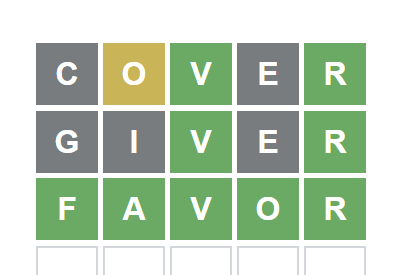}
         \caption{Guessing the word \textit{Favor}}       \label{fig:wordleFavor}
    \end{subfigure}
    \begin{subfigure}[!hbt]{0.24\textwidth}
         \centering      \includegraphics[width=\textwidth]{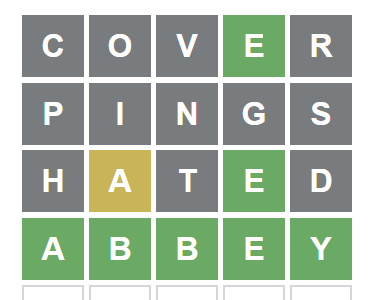}
         \caption{Guessing the word \textit{Abbey}}       \label{fig:wordleAbbey}
    \end{subfigure}
     
    \caption{Few Examples of \textit{Wordle} Solutions}
    \label{fig:wordle}
\end{figure}

The game accepts 12972 words as possible guesses for solutions while it has 2315 secret words as the actual solutions~\cite{anderson2022finding}. The tile based hints given to the player are as follows:
\begin{enumerate}
    \item \textbf{Green:} The entered character is in the expected solution and is in the expected position.
    \item \textbf{Yellow:} The entered character is in the expected solution but it is not in the expected position.
    \item \textbf{Gray:} The entered character is not in the expected solution.
\end{enumerate}
Figure~\ref{fig:wordle} shows four examples of \textit{wordle} solutions with the tile colours providing hints to the player to progress. Note from Fig~\ref{fig:wordleAbbey} and Fig~\ref{fig:wordleElder} that it is possible for the solution to have repeated characters. Also observe from Fig~\ref{fig:wordleElder} how these repeated characters are not clued in using the colour codes. The first line clues that \textit{E} is in the solution. But there is no indication as to how many times the character would appear. Another point to note is from Fig~\ref{fig:wordleFavor}, where it can be seen that \textit{Wordle} uses American spelling despite being developed by a person of UK origin and hosted at a \texttt{.co.uk} web address. Full automatic solving of \textit{Wordle} is a constraint satisfactory problem similar to that of the work by~\citet{de2013enabling}. However, creating such a solver is out of the scope of this study.

\section{Methodology}
\label{sec:method}
We obtain a manually prepared word list $G$ and derive the \textit{all word list}, $A$ as shown in Equation~\ref{eq:all} where, the $len(\cdot)$ function gives the length of the word $w$. This removes all words which are not of length 5 (e.g., \textit{cat}, \textit{dragons}). 

\begin{equation}
A=\underset{w \in G}{\cup}\begin{cases}
\{w\} & \text{if } len(w)=5 \\
\emptyset & \text{otherwise}  
\end{cases}
\label{eq:all}
\end{equation}

Next, we initialise the \textit{character frequency map}, $F$ of the format \verb|<character,value>| as shown in Equation~\ref{eq:charMapInit}, where $C$ is the alphabet.

\begin{equation}
    \underset{c \in C}{\forall}\bigg[F(c)=0 \bigg]
    \label{eq:charMapInit}
\end{equation}
 
Then, we populate the \textit{character frequency map}, $F$ as shown in Equation~\ref{eq:charMapPopu}, where $A$ is the all word list, $C$ is the alphabet, and $c_2$ is a character in the word $w$. 
 
\begin{equation}
\underset{w \in A}{\forall}\Bigg[\underset{c_1 \in C}{\forall}\Bigg[
F(c_1)=F(c_1)+\underset{c_2 \in w}{\Sigma}\begin{cases}
1 & \text{if } c_1 = c_2 \\
0 & \text{otherwise}
\end{cases}
\Bigg]\Bigg]
\label{eq:charMapPopu}
\end{equation}

Next, we update the \textit{character frequency map}, $F$, as shown in Equation~\ref{eq:charMap}, where $A$ is the \textit{all word list}, $C$ is the alphabet, and the $len(w)$ gives the character count (length) of the word $w$. This results in $F$ registering the global frequencies of each of the characters in the alphabet. 

\begin{equation}
\underset{c_1 \in C}{\forall}\Bigg[F(c_1)=\frac{F(c_1)}{\underset{w \in A}{\Sigma}len(w)}\Bigg]
\label{eq:charMap}
\end{equation}

We define the \textit{unique word list}, $W$ as shown in Equation~\ref{eq:uniq}, where $A$ is the \textit{all word list}, $c$ is a character in the word $w$, and the $len(\cdot)$ function gives the size of the set. This function makes sure that the $W$ only contains words that have five unique characters. This removes 5 letter words which have repeated characters (e.g., \textit{feels}). 

\begin{equation}
W= \underset{w \in A}{\cup}\begin{cases}
\{w\} & \text{if } len\big(\underset{c \in w}{\cup}\{c\}\big)=5 \\
\emptyset & \text{otherwise}  
\end{cases}
\label{eq:uniq}
\end{equation}

We define the \textit{word value map}, $M$ of the format \verb|<word,value>| as shown in Equation~\ref{eq:map}, where $W$ is the \textit{unique word list} from Equation~\ref{eq:uniq}, $c$ is a character in word $w$ and $F(c)$ is the value stored in the \textit{character frequency map} (created in Equation~\ref{eq:charMap}) for the character $c$.

\begin{equation}
\underset{w \in W}{\forall}\bigg\{
M(w)=\underset{c \in w}{\Sigma}F(c)   \bigg\}
\label{eq:map}
\end{equation}

Next, we define word overlap as shown in Equation~\ref{eq:over} where $w_1$ and $w_2$ are the candidate words and $i$ is a character. Thus, $I_{w_1,w_2}$ will carry the Boolean value \texttt{TRUE} is there is at least one common character between $w_1$ and $w_2$ or carry the Boolean value \texttt{FALSE} otherwise.

\begin{equation}
I_{w_1,w_2} = \exists i \text{ s.t } i \in w_1 \land i \in w_2
\label{eq:over}
\end{equation}

Next we define a greedy algorithm to select the current \textit{best words set} as shown in Equation~\ref{eq:best} where, $B$ is the set of best words, $B_0$ is the first element of $B$, and $M$ is a \textit{word value map} of the format \verb|<word,value>| (including but not limited to that which was defined in Equation~\ref{eq:map}). This process returns, as the result of $B(M)$, the highest valued words in $M$.

\begin{equation}
    B(M) = \underset{ w \in keys(M)}{\forall}\begin{cases}
    \{w\} & \text{if } B=\emptyset \text{ or } M(w) > M(B_0)\\
    B \cup \{w\} & \text{if } M(w) = M(B_0) \\
    B \cup \emptyset & \text{otherwise}
    \end{cases}
    \label{eq:best}
\end{equation}

We define the \textit{simplified best word list} as shown in Equation~\ref{eq:simplyBest} where $L$ is a list of words, $L_0$ is the first word in $L$, $\{L_1,\dots L_n \}$ is the list of words in $L$ other than $L_0$, and $n$ is the number of words in $L$. What this does is, given an $L$, it removes each of the words in $\{L_1,\dots L_n \}$ which has character overlaps with $L_0$ in an iterative manner and keeps the rest

\begin{equation}
S(L)=\begin{cases}
L & \text{if } n \leq 1\\
{L_0} \cup S\Bigg(\underset{ l \in \{L_1,\dots L_n \} }{\cup}\begin{cases}
\emptyset & \text{if } I_{L_0,l} \\
{l} & \text{otherwise}
\end{cases}\Bigg) & \text{otherwise}
\end{cases}
\label{eq:simplyBest}
\end{equation}

We define the \textit{filtered word value map}, $M^{'}$ of the format \verb|<word,value>| as shown in Equation~\ref{eq:fliter}, where $M$ is a \textit{word value map} of the format \verb|<word,value>| (including but not limited to that which was defined in Equation~\ref{eq:map}) and $w_1$ is a given filter word. This algorithm makes sure that $M^{'}_{M,w_1}$ contains the subset of \verb|<word,value>| pairs from $M$ such that, none of the keys have a character overlap with $w_1$. 

\begin{equation}
\underset{ w_2 \in keys(M)}{\forall}\bigg\{
M^{'}_{M,w_1}(w_2)=M(w_2) \text{ if } I_{w_1,w_2} = \text{ FALSE } \bigg\}
\label{eq:fliter}
\end{equation}

Finally we define candidate processing in Equation~\ref{eq:process}, where where $M$ is a \textit{word value map} of the format \verb|<word,value>| (including but not limited to that which was defined in Equation~\ref{eq:map}), the function $B(\cdot)$ is as defined in Equation~\ref{eq:best}, the function $S(\cdot)$ is as defined in Equation~\ref{eq:simplyBest},  the \textit{filtered word value map}, $M^{'}$ is as defined in Equation~\ref{eq:fliter}, and $w$ is a word.

\begin{equation}
P(M)=S(B(M)) \underset{ w \in S(B(M))}{\cup}\bigg[ P(M^{'}_{M,w})\bigg]
\label{eq:process}
\end{equation}

\section{Experiments and Results}
\label{sec:ExpRes}

For the manually prepared word list $G$ we used a publicly available word list from github\footURL{https://github.com/dwyl/english-words} which has over $466k$ English words. From the obtained $G$, using Equation~\ref{eq:all}, we derived the \textit{All word list}, $A$. We noted that $A$ contains only $21952$ words. Also, for the benefit of Equation~\ref{eq:charMap}, we calculated the total number of characters in the words in $A$ and observed that it has $109760$ characters. We show in Table~\ref{tab:chaFrq}, the character frequencies we calculated for the \textit{character frequency map}, $F$ as shown in Equation~\ref{eq:charMap}.

\begin{table}[!htb]
    \centering
    \caption{Calculated Character Frequencies}
    \label{tab:chaFrq}
    \begin{tabular}{|c|c|c|c|}
    \hline
     Character & Frequency & Character & Frequency \\
     \hline
        a & 0.1124 & n & 0.0546 \\
        b & 0.0268 & o & 0.0653 \\
        c & 0.0330 & p & 0.0263\\
        d & 0.0355 & q & 0.0015\\
        e & 0.0994 & r & 0.0648\\
        f & 0.0147 & s & 0.0754\\
        g & 0.0236 & t & 0.0491\\
        h & 0.0290 & u & 0.0399\\
        i & 0.0661 & v & 0.0114\\
        j & 0.0057 & w & 0.0135\\
        k & 0.0224 & x & 0.0042\\
        l & 0.0556 & y & 0.0314\\
        m & 0.0318 & z & 0.0069\\
        \hline
    \end{tabular}
\end{table}

\begin{figure}[!htb]
    \centering
    \includegraphics[width=0.35\textwidth]{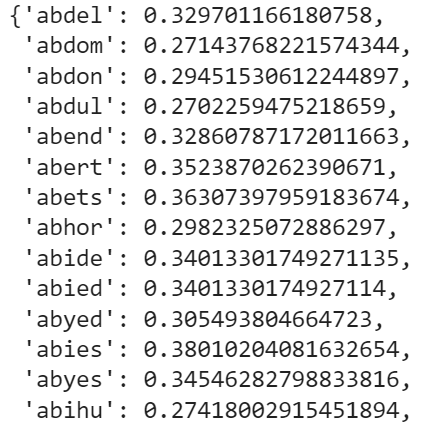}
    \caption{A part of the \textit{Word Value Map}}
    \label{fig:wordFrequs}
\end{figure}

An interesting observation we can make from Table~\ref{tab:chaFrq} is that the character \textit{a} is more frequent than character \textit{e}. This contrasts the character frequency behavious reported in earlier literature~\cite{KeatingFrequency,WallEnglish}. Other than that, all the other character frequencies aligns with common wisdom. 
After using Equation~\ref{eq:uniq} to obtain $W$, we observed to have left with $13672$ words. This is $62.28\%$ of the word count we had in $A$. It is an interesting to observe that in the 5 letter word domain, majority of the words seem to have 5 unique characters rather than having repeated characters. Figure~\ref{fig:wordFrequs} shows a part of the \textit{word value map}, $M$ of the format \verb|<word,value>| as generated in Equation~\ref{eq:map}.

When we executed Equation~\ref{eq:process}, and extracted our word suggestions. At this point, we faced with a problem. It was the fact that some of the words that were suggested as candidates were not being accepted as valid words by \textit{Wordle}. This, we observe, is due to the fact that our word list is richer than that used in \textit{Wordle}. Because of this, we had to manually drop a number of high ranking words. The words that we had to drop are: \textit{aires}, \textit{erisa}, \textit{saire}, \textit{luton}, \textit{tould}, \textit{unold}, \textit{dunlo}, \textit{xdmcp}, and \textit{aries}.

\begin{figure}[!hbt]
     \begin{subfigure}[!hbt]{0.24\textwidth}
         \centering      \includegraphics[width=\textwidth]{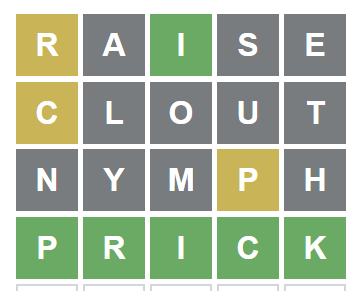}
         \caption{Guessing the word \textit{Prick}}       \label{fig:wordlePrick}
    \end{subfigure}
    \begin{subfigure}[!hbt]{0.24\textwidth}
         \centering      \includegraphics[width=\textwidth]{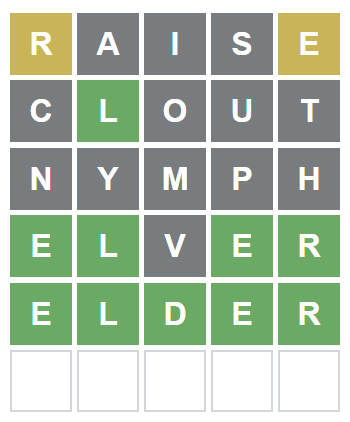}
         \caption{Guessing the word \textit{Elder}}       \label{fig:wordleElder}
    \end{subfigure}
    \caption{Two Examples of \textit{Wordle} Solutions with the selected words \texttt{['raise', 'clout', 'nymph']} }
    \label{fig:wordleNew}
\end{figure}

The highest ranking suggestion to be accepted by Wordle was \texttt{['serai', 'nould']}. However, this had to be abandoned due to having only two words and thus only covering 10 characters. The highest ranked 3 word set accepted by wordle was \texttt{['aesir', 'donut', 'lymph']}. Logically, these are the best three words to use. However \texttt{aesir} is not a very common word used by our alpha testers. Thus, we opted to settle with the next best word list  \texttt{['raise', 'clout', 'nymph']}. Figure~\ref{fig:wordleNew} shows two examples of solved \textit{Wordle}s with the selected words.

\section{Conclusion}
\label{sec:Conlu}
The objective of this work was~\obj. We succeeded in that target by discovering the words \texttt{['raise', 'clout', 'nymph']} to be the optimum starting words. At this point it should be noted that the auto solver proposed by~\citet{Neil2022Ruining} as reported by~\citet{Groux2022Best} also suggests \texttt{raise} as a good starting word. This overlap with our result proves that our methodology is sound. 

\bibliographystyle{IEEEtranN}
\footnotesize
\bibliography{ref}
\normalsize

\end{document}